\documentclass[10pt,twocolumn,letterpaper]{article}

\usepackage{cvpr}              

\usepackage{graphicx}
\usepackage{amsmath}
\usepackage{amssymb}
\usepackage{booktabs}
\usepackage{multirow}
\usepackage{comment}
\usepackage[accsupp]{axessibility}
\usepackage[pagebackref,breaklinks,colorlinks]{hyperref}
\renewcommand\footnotemark{}

\usepackage[capitalize]{cleveref}
\crefname{section}{Sec.}{Secs.}
\Crefname{section}{Section}{Sections}
\Crefname{table}{Table}{Tables}
\crefname{table}{Tab.}{Tabs.}

\def\cvprPaperID{5806} 
\def\confName{CVPR}
\def\confYear{2024}

\begin{document}

\title{ShapeMatcher: Self-Supervised Joint Shape Canonicalization, \\ Segmentation, Retrieval and Deformation}

\author{Yan Di$^{*}$, Chenyangguang Zhang$^{*}$, Chaowei Wang$^{*}$, Ruida Zhang, Guangyao Zhai, 
Yanyan Li,\\
Bowen Fu,
Xiangyang Ji and Shan Gao\\
\textsuperscript{}Tsinghua University, \textsuperscript{}Northwestern Polytechnical University, \\
\tt\small{\{zcyg22, zhangrd21\}@mails.tsinghua.edu.cn}
\\
\thanks{*Authors with equal contributions.}
}
\maketitle

\begin{abstract}
In this paper, we present ShapeMatcher, a unified self-supervised learning framework for joint shape canonicalization, segmentation, retrieval and deformation.
Given a partially-observed object in an arbitrary pose, we first canonicalize the object by extracting point-wise affine-invariant features, disentangling inherent structure of the object with its pose and size.
These learned features are then leveraged to predict semantically consistent part segmentation and corresponding part centers.
Next, our lightweight retrieval module aggregates the features within each part as its retrieval token and compare all the tokens with source shapes from a pre-established database to identify the most geometrically similar shape.
Finally, we deform the retrieved shape in the deformation module to tightly fit the input object by harnessing part center guided neural cage deformation.
The key insight of ShapeMaker is the simultaneous training of the four highly-associated processes: canonicalization, segmentation, retrieval, and deformation, leveraging cross-task consistency losses for mutual supervision.
Extensive experiments on synthetic datasets PartNet, ComplementMe, and real-world dataset Scan2CAD demonstrate that ShapeMatcher surpasses competitors by a large margin.
Code is released at \url{https://github.com/Det1999/ShapeMaker}.
\end{abstract}

\section{Introduction}
In recent years, there has been a notable surge in research interest focused on generating high-quality 3D models from scans of complex scenes~\cite{AUic20,MPic19,MRic19,MScv22,NRcv21,NScv22,PLcv22}.
This technology encourages extensive applications in both artistic creation~\cite{DAec20,JLcv21}, robotics~\cite{zhai2023monograspnet,zhai2023sg} and 3D scene perception~\cite{T3cv20,H3cv21}.
Existing methods~\cite{WDcv19,T3cv20} typically directly utilize deep neural networks to reconstruct 3D models from imperfect scans. 
However, the presence of noise and occlusions poses a significant challenge in accurately capturing fine-grained geometric structures.
To overcome this, Retrieval and Deformation ($\mathbf{R\&D}$) techniques~\cite{JEic19,JLcv21,ASto12,DAec20,NCcv20,3Dcv19,SFni20,ROto17} have been developed. 
These methods generally involve two key steps: first, identifying the most geometrically similar source shape from a pre-curated 3D database; and second, deforming the retrieved shape to achieve precise alignment with the target input.
The $\mathbf{R\&D}$ approach is particularly effective in producing 3D models enriched with fine details from source shapes.

\begin{figure}[t]
   \centering
   \includegraphics[width=\linewidth]{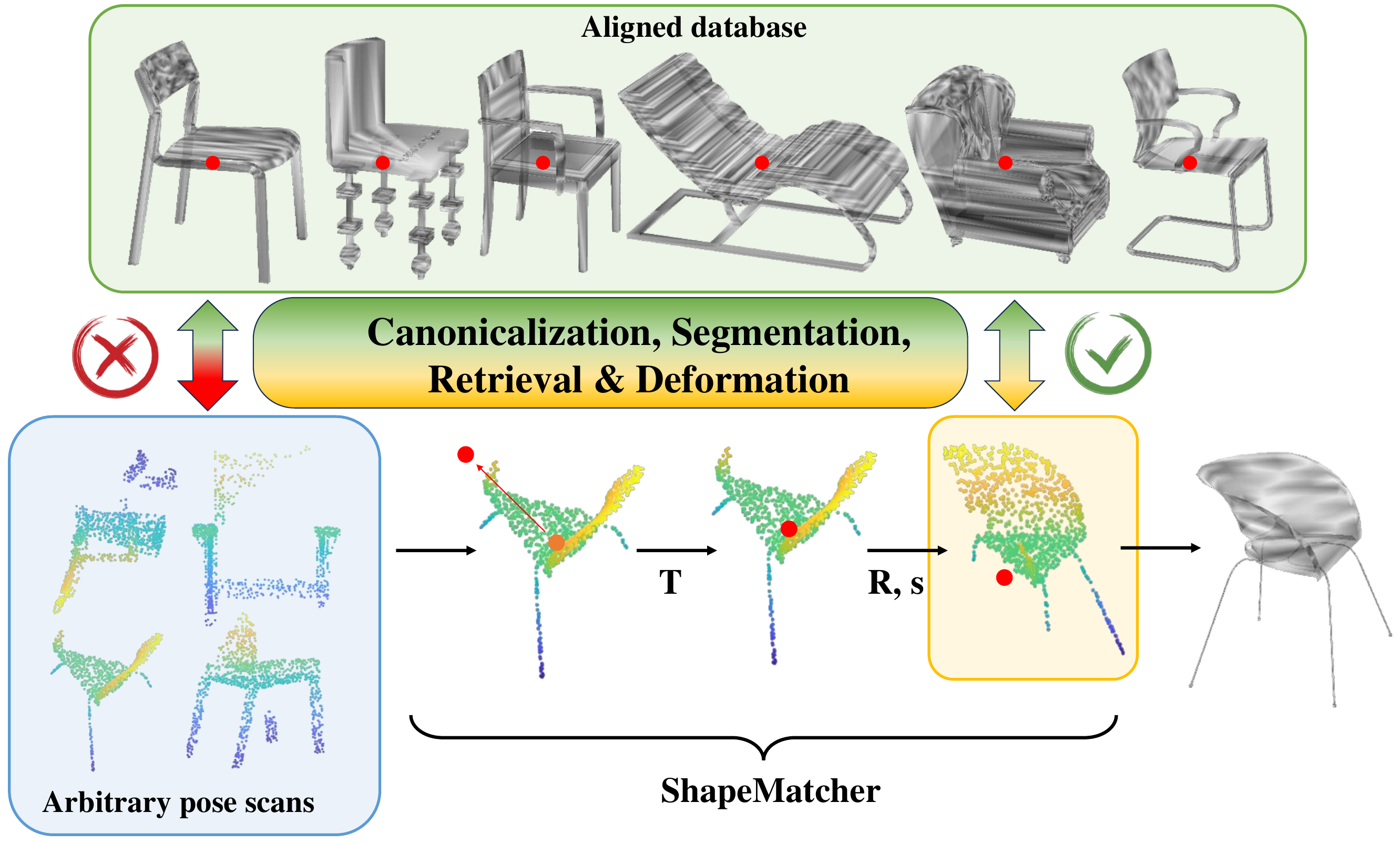}
    \vspace{-0.2cm}
   \caption{\textbf{Illustration of ShapeMatcher.} 
   Objects obtained from real-world scans are typically noisy, partial and exhibit various poses, making it challenging to conduct an effective $R\&D$ process (Red '\textcolor{red}{X}' on the left). 
   To address this issue, we propose ShapeMatcher that first canonicalizes the objects and then segments them into semantic parts, facilitating $\mathbf{R\&D}$ processes (Green '\textcolor{green}{$\checkmark$}' on the right).
   }
   \label{fig:pre}
\vspace{-0.5cm}
 \end{figure}

However, existing $\mathbf{R\&D}$ methods usually encounter two primary challenges that make them susceptible to noise, occlusion and pose variations, and difficult to be practically utilized. 
\textbf{1)} Most $\mathbf{R\&D}$ techniques~\cite{JEic19,JLcv21,ASto12,DAec20,NCcv20,3Dcv19,SFni20,ROto17,URED} operate under the assumption that target shapes are aligned in a pre-processed canonical space. 
Typically, these methods are trained and tested on datasets where shapes have been manually adjusted to this canonical state.
However, when these methods are deployed in real-world settings, they necessitate either manual alignment of scanned objects or the use of additional pose estimation networks~\cite{di2022gpv, zhang2022rbp, zhang2022ssp, su2023opa, di2021so}.
Such procedures are not only time-intensive and laborious but also prone to yielding inconsistent results.
This limitation significantly impedes the direct application of these methods in real-world scenarios.
\textbf{2)} Previous methods~\cite{JLcv21,DAec20} do not design specially for dealing with partially-observed shapes, making it difficult to handle occluded objects. 
Although U-RED~\cite{URED} considers the partial target shapes as input in the $\mathbf{R\&D}$ process, it directly encodes the shape as a global embedding, which is not robust when dealing with significant occlusion.

\begin{figure*}
   \centering
   \includegraphics[width=0.95\linewidth]{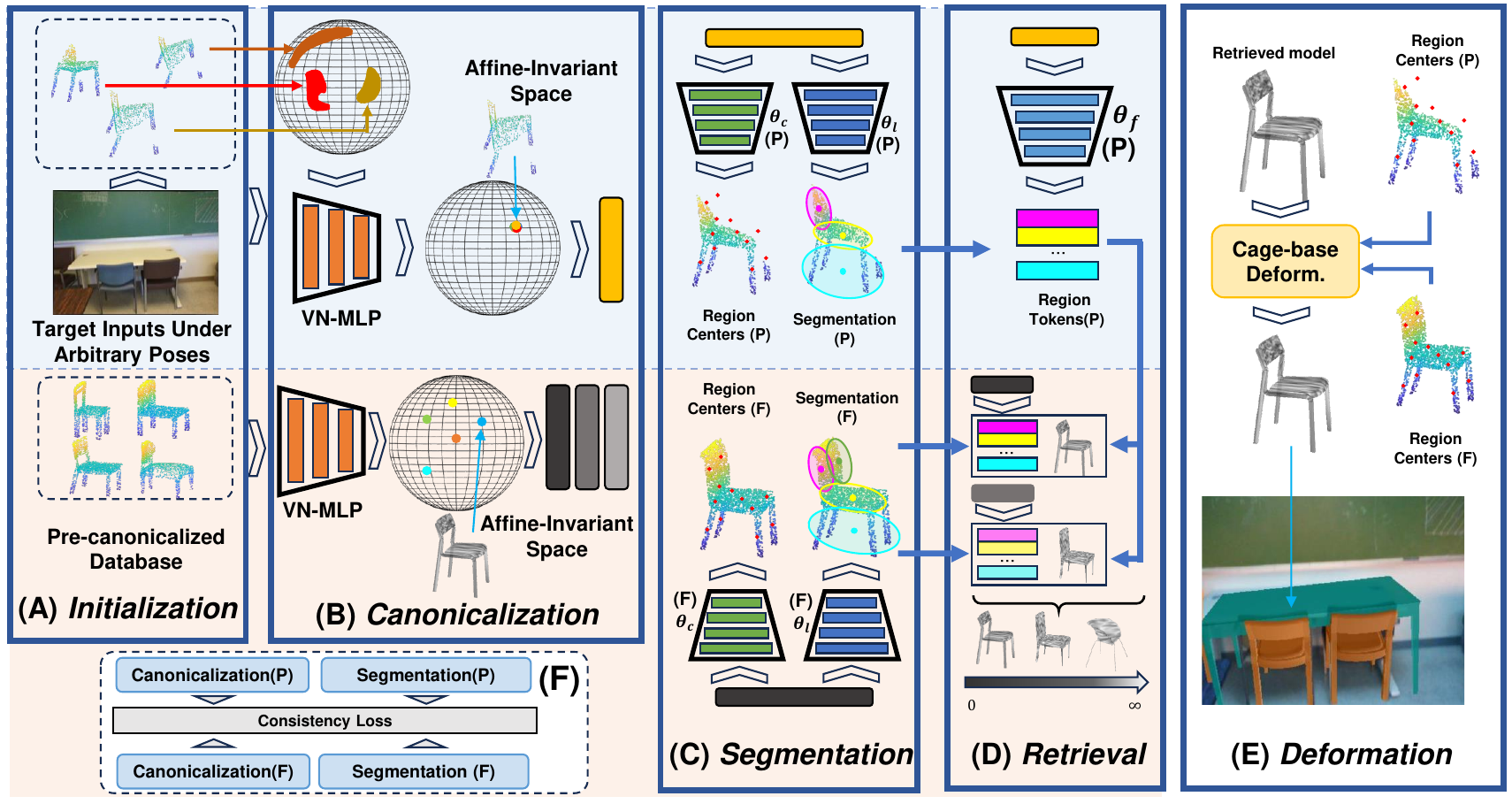}
   \caption{\textbf{The pipeline of ShapeMatcher.} 
   Given a target point cloud obtained from a single-view scan and a pre-established database (A), ShapeMatcher generates the fine-grained reconstruction result using the joint 4 modules including \textit{Canonicalization} (B), \textit{Segmentation} (C), \textit{Retrieval} (D) and \textit{Deformation} (E), where the first three contains the partial branch for target processing and the full branch for source processing.
   Specifically, the target and source inputs are first canonicalized into the same affine-invariant space (B).
   Then, the semantic-consistent region segmentation is yielded from the affine-invariant features (C).
   The segmented regions are fed to the region-weight retrieval module (C) and the part center guided neural cage deformation module (E) for occlusion-robust $\mathbf{R\&D}$ process.
   During training, the partial-full consistency losses (F) are enforced for the two branches.
   }
  \label{fig:mid1}
  \vspace{-0.5cm}
 \end{figure*}

To address the aforementioned two challenges, in this paper, we present ShapeMatcher, a novel framework that extends traditional $\mathbf{R\&D}$ pipeline to joint self-supervised learning of object canonicalization, segmentation, retrieval and deformation.
Our core contribution lies in that the four highly-associated processes can be trained simultaneously and supervise each other via constructing several cross-task consistency losses (Fig.~\ref{fig:pre}).
Specifically, given a partially-observed object scan in an arbitrary pose, ShapeMatcher processes the objects in four steps.
First, we follow~\cite{katzir2022vnt}, which is based on Vector Neurons~\cite{VNic21}, to extract SE(3)-invariant point-wise features by progressively separating translation and rotation.
We further follow~\cite{chen20223gnt} to normalize the features to disassociate the object scale.
Until here, we successfully obtain affine-invariant point-wise features by disentangling object's inherent structure with its pose and size.
This facilitates the \textit{Canonicalization} of the observed object based on these intrinsic characteristics (Fig.~\ref{fig:mid1} (B)).
Then we predict semantically consistent part segmentation and corresponding part centers by feeding the learned features into our \textit{Segmentation} module (Fig.~\ref{fig:mid1} (C)). 
Based on the part segmentation, in the \textit{Retrieval} module (Fig.~\ref{fig:mid1} (D)), we aggregate features within each part and collect them together as a comprehensive retrieval token of the object.
For partial objects, we introduce a region-weighted strategy, which assigns a weight to each part according to the point inside it.
Parts with more points are assigned higher weights during retrieval, which is proved to be robust to occlusions.
We compare the tokens of the target object with each shape in the pre-constructed database to identify the most geometrically similar (most similar tokens) source shape.
In the final \textit{Deformation} module (Fig.~\ref{fig:mid1} (E)), the retrieved source shape is deformed to tightly match the target object via part center guided neural cage deformation~\cite{NCcv20}. 

To summarize, our main contributions are:
\begin{itemize}
\item We introduce ShapeMatcher, a novel self-supervised framework for joint shape canonicalization, segmentation, retrieval and deformation, handling partial target inputs under arbitrary poses. 
Extensive experiments on the synthetic and real-world datasets demonstrate that ShapeMatcher surpasses existing state-of-the-art approaches by a large margin.
\item We demonstrate that the four highly-associated tasks: canonicalization, segmentation, retrieval and deformation, can be effectively trained simultaneously and supervise each other via constructing consistencies. 
\item We develop the region-weighted retrieval method to mitigate the impact of occlusions in the $\mathbf{R\&D}$ process.
\end{itemize}
\section{Related Works}
\textbf{Neural Shape Representation.}
The compact representation of 3D shapes in latent space, based on deep learning, has been a focal point for many researchers. 
Some attempts, such as \cite{LIcv19,ONcv19,DScv19,MSni20,NERF,PNcv22,CNcv21, zhang2023ddf}, employ neural networks to construct an implicit function, while others \cite{LRic18,PFic19,STar19,PGca20, zhai2023commonscenes} directly model the shape of objects explicitly using generative models. 
Another common architecture in 3D shape representation learning, as seen in \cite{SCcv17,PCni22,MAto23,SVcv21,P2ic19}, is to use an encoder-decoder approach to generate latent representation vectors for various shapes. 
Although these methods have demonstrated impressive representation performance, they often struggle to generate fine-grained shapes when dealing with occlusion and noise. 

\textbf{CAD Model Retrieval and Deformation.}
Retrieval and Deformation ($\mathbf{R\&D}$) methods lead another way to recover fine-grained geometric structures.
Previous works directly retrieve the most similar CAD models by comparing the similarity of expression vectors in either descriptor space \cite{ROto17,AMai08} or the latent space of neural networks \cite{JEto15,JEic19,ROCA,ETic19}. 
Considering the subsequent deformation error, recent efforts introduce deformation-aware embeddings \cite{DAec20} or proposed new optimization objectives \cite{CDec20} to better capture the fine structure of deformed target objects. 
Nevertheless, these methods yield deteriorated performance when facing partial and pose-agnostic target shapes in real world. 
\cite{URED} achieves an one-to-many retrieval module for addressing the issue caused by partial observations, however, it receives canonicalized target shapes as input, which limit its applicability facing pose-agnostic target shapes in real world.
As the retrieved models often exhibit some deviation from the target shape, the deformation module is used to minimize this discrepancy. 
Traditional approaches \cite{ARso07,NRcg08,PGic18} aim to fit the target shape by directly optimizing the deformed shape. 
Neural network based techniques attempt to learn a set of deformation priors from a database of models. 
They represent deformations as volume warping \cite{DFwc18,LFac19}, cage deformations \cite{NCcv20}, vertex offsets \cite{3Dcv19}, or flows \cite{SFni20, zaccaria2023self}. 
These methods typically constrain two shapes are aligned in the same coordinate system, making them challenging to apply in real-world scenarios.

\textbf{SO(3)-Equivariant Methods.}
An increasing body of work \cite{TFar18,CGar18,LSec18,3Sni18,CMni19} has initiated research on SO(3) equivariance. 
These efforts are mostly based on steerable convolutional kernels \cite{AWar20}. 
On the other hand, another set of works achieves equivariance through pose estimation. 
\cite{PNcv17} estimates the object's pose to factor out SO(3) transformations, achieving approximate equivariance. 
While \cite{CCar21} learns pose estimation in a fully unsupervised manner, the equivariant backbone they employ \cite{ACcv20} achieves equivariance primarily through data augmentation, leading to limited generalization.
In this paper, we employ Vector Neural Multi-Layer Perceptron \cite{VNic21} as the backbone to get neural invariant features for object canonicalization. 
It achieves SO(3) equivariance by lifting traditional scalar neurons to vector neurons.

\section{Method}

\textbf{Overview.}
ShapeMatcher consists of 4 modules, corresponding to the 4 highly-associated tasks.
Each of the first 3 modules: \textit{Canonicalization, Segmentation} and \textit{Retrieval} modules have two parallel branches, one for complete point cloud (in orange background) and the other for partial object input (in blue background). 
As shown in Fig.~\ref{fig:mid1}, given a partial target shape $S_{tgt} \in \mathbb{R}^{N \times 3}$ in an arbitrary pose, in the \textit{Canonicalization} module, we progressively decouples its inherent shape with rotation $R_{tgt} \in SO(3)$, translation $T_{tgt} \in \mathbb{R}^3$ and the $3 \mathrm{D}$ metric size $s_{tgt} \in \mathbb{R}^3$ via VN-MLP~\cite{VNic21, katzir2022vnt,chen20223gnt}, yielding the affine-invariant point-wise features $F_{tgt}$.
The object can then be canonicalized via inverse transformation based on intrinsics $\{R_{tgt}, T_{tgt}, s_{tgt}\}$.
In the \textit{Segmentation} module, $F_{tgt}$ is fed into a 4-layer MLP network to predict $M$ parts and corresponding part centers $\{K_{tgt}^1,K_{tgt}^2,...,K_{tgt}^M\}$.
The segmentation is semantically consistent across each category and thus can be matched and compared for $\mathbf{R\&D}$.
In the \textit{Retrieval} module, inside each region $M^i$, we aggregate the features of all points inside it as its retrieval token $Q^i$.
The retrieval token for the object is then represented as $\mathbf{Q_{tgt}}=\{Q_{tgt}^1,Q_{tgt}^2,...,Q_{tgt}^M\}$.
Similarly, during training, we obtain the intrinsics $\{R_{src}, T_{src}, s_{src}\}$, part centers $\{K_{src}^1,K_{src}^2,...,K_{src}^M\}$ and retrieval tokens $\mathbf{Q_{src}}=\{Q_{src}^1,Q_{src}^2,...,Q_{src}^M\}$ via the branch for complete point cloud.
By comparing $\mathbf{Q_{tgt}}$ and $\mathbf{Q_{src}}$ of each source shape inside the database, we identify the most geometrically similar source shape $S_r$.
In the final $\textit{Deformation}$ module, $K_{tgt}$ and $K_{src}$ are leveraged to guide the neural cage deformation~\cite{NCcv20} to deform the retrieved $S_r$ towards $S_{tgt}$, yileding $S_{src}^{dfm}$.

\subsection{Canonicalization}\label{can}
As shown in Fig.~\ref{fig:mid1} (B), the \textit{Canonicalization} module takes the target point cloud $S_{tgt}$ as input and disentangle the inherent structure of $S_{tgt}$ with the intrinsics $\{R_{tgt}, T_{tgt}, s_{tgt}\}$, yielding a point-wise affine-invariant feature $F_{tgt}$. 
Specifically, we follow VN-MLP~\cite{VNic21, katzir2022vnt,chen20223gnt} to first decouple translation via VNT~\cite{katzir2022vnt} and then extract rotation via VNN~\cite{VNic21, katzir2022vnt}.
We further follow ~\cite{chen20223gnt} to normalize the SE(3)-invariant features obtained above the remove the influence of scaling, yielding $F_{tgt}$ as follows,
\begin{equation}
  R_{tgt}, T_{tgt}, F_{tgt}^{*} = \mathbf{VN\mbox{-}MLP}(S_{tgt})
  \label{eq:7}
\end{equation}
\begin{equation}
  s_{tgt}, F_{tgt} = normalize(F_{tgt}^{*})
  \label{eq:8}
\end{equation}
where $F_{tgt}^{*}$ denotes the SE(3)-invariant features and $F_{tgt}$ denotes the affine-invariant features.
Thereby, the object can be canonicalized with intrinsics as,
\begin{equation}
  S^{c}_{tgt} = s_{tgt}R_{tgt}S_{tgt} + T_{tgt}
  \label{eq:9}
\end{equation}
where $S^{c}_{tgt}$ denotes the normalized and canonicalized shape of $S^{tgt}$. 
During training, in order to ensure that $F_{tgt}$ fully encapsulate the geometric information of $S_{tgt}$, we integrate a supplementary reconstruction branch which takes $F_{tgt}^{*}$ as input and reconstruct $S^{c}_{tgt}$ in the affine-invariant space~\cite{katzir2022vnt}.
Please refer to the Supplementary Material for details.
For source shape $S_{src}$ from the database, we follow the same procedures to extract $F_{src}$.

\subsection{Segmentation}
Given the affine-invariant features $F_{tgt}$, we segment the input point cloud $S_{tgt}$ into $M$ semantically consistent parts.
We use a 4-layer MLP $\Theta_{l}$ (Fig.~\ref{fig:mid1} (C)) to predict a one-hot segmentation label for each point and use another 4-layer MLP $\Theta_{c}$ to predict $M$ part centers $\{K_{src}^1,K_{src}^2,...,K_{src}^M\}$.
Noteworthy, we don't need any ground truth annotations in this segmentation process.
Our experiments show that the network can automatically learn semantically consistent segmentation solely through consistency supervision from the other three tasks.
For source shape $S_{src}$, we follow a similar process to obtain $\{K_{src}^1,K_{src}^2,...,K_{src}^M\}$.

\subsection{Retrieval}
The retrieval network aims to identify the model $S_{src}$ from an existing database that bears the closest resemblance to the target object $S_{tgt}$ after deformation. 
Traditional methods \cite{URED,JLcv21} directly extract the global features of objects for retrieval, which typically struggles with heavy occlusion since the global features are susceptible to noise and occlusion, and prone to producing erroneous retrieval results. 
In contrast, we employ a novel region-weighted retrieval method to explicitly encode independent and semantically consistent regions of the shape. 
This allows us to accurately handle partial shapes by identifying the visible regions to retrieve models most similar to the target.

Specifically, the part segmentation network $\Theta_{l}$ takes $F_{tgt}$ as input to predict $M$ regions $F_{seg} = [C_1, C_2, C_3, ..., C_N]^\top$ of $S^{c}_{tgt}$, where $F_{seg} \in \mathbb{R}^{N\times M }$, $C_i\in \mathbb{R}^M$ represents the probability of point $i$ belonging to each part center.
Then we use another 4-layer feature aggregator $\Theta_f$ (Fig.~\ref{fig:mid1} (D)) to extract the retrieval tokens $Q$ of all parts as follows,
\begin{equation}
  F_{cls} = F_{seg}^\top * \Theta_f(F_{tgt}),
  \label{eq:r1}
\end{equation}
\begin{equation}
  Q_{tgt} = F_{cls} / (\sum_{n=1}^NF_{seg}^{(n)}).
  \label{eq:r2}
\end{equation}
where $Q_{tgt} \in \mathbb{R}^{M \times C}$ contains the $C$-dimensional retrieval tokens for all the $M$ regions.
Here we employ a soft assignment strategy where each point inside $S_{tgt}$ is estimated $M$ values describing the probabilities belonging to each of the $M$ parts.
Therefore, we first aggregate features $F_{cls}$ on all points belonging to each part and then normalize $F_{cls}$ using the sum of probabilities of points in each part, as in Eq.~\ref{eq:r1} and Eq.~\ref{eq:r2}.
Following a similar strategy, we can obtain $Q_{src}$ for each source model in the pre-curated database.
We just need to compare $Q_{tgt}$ with the retrieval tokens $Q_{src}$ of all source shapes using the weighted $\mathcal{L}_1$ distance,
\begin{equation}
    Dis = \omega \sum\mathcal{L}_1(Q_{tgt} - Q_{src})
\end{equation}
where vector $\omega \in \mathbb{R}^{1 \times M}$ stores the ratio of point number of each part with respect to the total point number $N$.
Intuitively, parts with smaller point numbers contribute less in calculating the distance score, which reduces the influence of noise and occlusion.
The source shape $S_r$ with the smallest distance score is identified as the best retrieval.

\subsection{Deformation}
The \textit{Deformation} module aims to deform the retrieved shape $S_r$ to tightly match the target shape $S_{tgt}$.
We utilize the neural cage scaffolding strategy as in~\cite{NCcv20,KPcv21}. 
First, the neural cages $C_{src}$ for $S_r$ is pre-calculated.
We utilize the part centers ($K_{tgt}$ and $K_{src}$) to control the vertice offsets $C_{src2tgt}$ of the neural cage $C_{src}$ to match $S_{tgt}$.
In particular, we employ a neural network $\Theta_I$ to predict an influence vector $I \in \mathbb{R}^{N_c \times M}$ for each point concerning all cage vertices by $I = \Theta_I (concat(F_{tgt},F_{src}))$, where $N_c$ denotes the number of vertices used in $C_{src}$. 
$C_{src2tgt}$ is computed through the influence vectors $I$ and the differences between region centers (Fig.~\ref{fig:mid1} (E)):
\begin{equation}
   C_{src2tgt} = C_{src} + \sum_{i=1}^M I_i(K_{src}^{(i)}-K_{tgt}^{(i)}),
  \label{eq:defo2}
\end{equation}
Finally, we employ a sparse cage scaffolding strategy~\cite{NCcv20,KPcv21} to achieve the deformation field  of $S_{src}$. 
The deformed shape $S_{src2tgt}$ of $S_{src}$ can be expressed as follows:
\begin{equation}
   S_{src2tgt} = S_{src} + \Psi(C_{src},C_{src2tgt}),
  \label{eq:defo}
\end{equation}
where $\Psi$ computes the displacement of each point in $S_{src}$ by evaluating the differences between $C_{src}$ and $C_{src2tgt}$, thereby achieving deformation. 

\subsection{ShapeMatcher: Joint Training}\label{training}
Our core insight in ShapeMatcher is that the four highly-associated tasks: \textit{Canonicalization, Segmentation, Retrieval} and \textit{Deformation} can be trained simultaneously and supervise each other via introducing cross-task consistency terms.
We mainly introduce two types of losses here, \textit{i.e.} partial-full consistency losses and task-oriented loss.
For more details, please refer to the Supplementary Material.

\noindent \textbf{Task-Oriented Loss.}
In the \textit{Canonicalization}, we mainly use Chamfer Distance to constrain the canonicalized $S^{c}_{tgt}$ and $\hat{S^{c}}_{tgt}$ predicted in the affine-invariant space by the supplementary reconstruction branch, to enforce the affine-invarianty of $F_{tgt}^{*}$ in Sec. \ref{can}, by
\begin{equation}
    \mathcal{L}_{can} = dis_{cham}(S^{c}_{tgt},\hat{S^{c}}_{tgt})+orth(R_{tgt}),
\end{equation}
with $orth(R_{tgt})$ serving the purpose of enforcing the orthogonality of matrices.

In the \textit{Segmentation}, to keep consistency between the part segmentation and the predicted part center, we jointly train $\Theta_l$ and $\Theta_c$ with the following loss, which enforce that each predicted part center $K^{i}_{tgt}$ approximately lies in the center of all points belonging to the part $M_i$,
\begin{equation}
    \mathcal{L}_{seg} = \sum_{m=1}^{M} \Vert K_{tgt}^{(m)} - (F_{seg}^\top * S^{c}_{tgt})^{(m)} \Vert_2
\end{equation}

To train the \textit{Retrieval} and \textit{Deformation} simultaneously, for an input target $S_{tgt}$, we randomly select a source model $S_{src}$ from the database for training. 
Specifically, to eliminate the influence of occlusion, we do not directly use the global Chamfer Distance of $S_{tgt}$ and $S_{src}$ as ground truth.  
Instead, we employ a regional supervision strategy, ensuring that occluded areas do not contribute to the training of retrieval network. 
Taking the $i$-th region as an example, $S_{tgt}^i$ represents all points in $S_{tgt}$ that belong to the $i$-th region. 
We calculate the average of the nearest distances $D_i$ from each point in $S_{tgt}^i$ to the deformed shape $S_{src}^{dfm}$ to enforce the learning of the regional retrieval tokens by
\begin{equation}
  \mathcal{L}_{retrieval} = \frac{1}{M}\sum_{i=1}^{M}MSE(Q_{tgt}^{(i)}-Q_{src}^{(i)},D_i).
  \label{eq:rloss}
\end{equation}
The deformation loss is achieved by directly constraining the Chamfer Distance between $S_{tgt}$ and $S_{src}^{dfm}$, expressed as:
\begin{equation}
  \mathcal{L}_{deform} = dis_{cham}(S_{tgt},S_{src}^{dfm}) + \Vert I \Vert_2
  \label{eq:dloss}
\end{equation}
where we regularize $I$ using the L2 norm.

\noindent \textbf{Partial-Full Consistency Losses.}
In the first two modules: \textit{Canonicalization, Segmentation}, the full branch serves as a guidance to enhance the learning of the partial branch.
Therefore, in each module, we can enforce corresponding consistency terms between the outputs of the two parallel branches (Fig.~\ref{fig:mid1} (F)).

During the consistency training process, for randomly selected full input $S_{full}$, we generate a mask $U_{f2p} \in \mathbb{R}^N$ to crop it to simulate the situation of a partial input $S_{partial} = S_{full} U_{f2p}$.

In the \textit{Canonicalization} module, to enforce the consistency in the affine-invariant space between the two branches, we apply the same transformation $U_{f2p}$ to $S_{full}^{c}$ in the affine-invariant space as before and then use the chamfer distance to constrain its distance to $S_{partial}^{c}$:
\begin{equation}
  \mathcal{L}_{ccan} = dis_{cham}(S^{c}_{partial},S^{c}_{full} U_{f2p})
  \label{eq:rrloss}
\end{equation}
Similarly, in the \textit{Segmentation} module, for the consistency constraint of the part center prediction network $\Theta_{c}$, we directly use the Chamfer Distance to constrain the region centers detected by the two branches: 
\begin{equation}
  \mathcal{L}_{ccen} = dis_{cham}(K_{partial},K_{full}).
  \label{eq:kloss}
\end{equation}
In the segmentation network $\Theta_{l}$, we mask the segmentation results of the full branch $F_{seg}^{(full)}$ and compare them with the results of the partial branch $F_{seg}^{(partial)}$:
\begin{equation}
  \mathcal{L}_{cseg} = dis_{cham}(F_{seg}^{(full)} U_{f2p},F_{seg}^{(partial)}).
  \label{eq:seloss}
\end{equation}

\textbf{Joint Training.}
Generally, the joint training of ShapeMatcher is divided into three stages.
First, we train the full branch by $\mathcal{L}_{can}$ and $\mathcal{L}_{seg}$ for construction of \textit{Canonicalization} and \textit{Segmentation} ability.
Second, the partial branch is introduced and trained by both the task-oriented losses for \textit{Canonicalization} and \textit{Segmentation} $\mathcal{L}_{can}$ and $\mathcal{L}_{seg}$ and the partial-full consistency loss terms $\mathcal{L}_{ccan}$, $\mathcal{L}_{ccen}$ and $\mathcal{L}_{cseg}$.
Finally, after training \textit{Canonicalization} and \textit{Segmentation} of the both branches, $\mathcal{L}_{retrieval}$ and $\mathcal{L}_{deform}$ are adopted for joint \textit{Retrieval} and \textit{Deformation} training simultaneously utilizing the both branches to handle partial target inputs and full source inputs respectively. 

\section{Experiments}

\subsection{Experimental Setup}
In this section, we mainly focus on $\mathbf{R\&D}$ experiments, which better reflects the overall performance of the system.
The ablations and analysis also demonstrate the effectiveness of considering joint \textit{Canonicalization} and \textit{Part Segmentation}.

\textbf{Datasets.} 
We evaluate the effectiveness of our joint framework using three datasets: two synthetic datasets, PartNet \cite{PNcv19} and ComplementMe \cite{CMto17}, and one real-world dataset, Scan2CAD \cite{S2C}. 
For datasets PartNet and ComplementMe, we follow the same database splits as in \cite{JLcv21}, separating their target inputs into training and testing sets. 
In our training process, we exclusively employ mesh models and do not utilize part segmentations as in \cite{JLcv21}, since the process of ShapeMatcher is fully self-supervised and does not need any additional annotations.
The shapes used in PartNet and ComplementMe datasets are sourced from ShapeNet \cite{Shape}. 
PartNet comprises 1,419 source models in the database, with 11,433 target models in the training set and 2,861 in the testing set.
In ComplementMe, the numbers are 400, 11,311 and 2,825 respectively. 
In the synthetic cases, three categories of tables, chairs, and cabinets are evaluated on both datasets.
Scan2CAD \cite{S2C} is a real-world dataset developed based on ScanNet \cite{SCANN} with capacity of 14,225 objects. 
The input point cloud data on Scan2CAD is generated by reverse-projecting the depth images. 
In the real-world cases, we conduct training on the categories of tables, chairs, and cabinets from PartNet and directly testing on Scan2CAD.

\textbf{Baselines.}
Both baseline methods, Uy \textit{et al}. \cite{JLcv21} and U-RED \cite{URED}, are trained using the same data partitioning strategy stated above. 
To ensure fairness in comparison with ShapeMatcher, we augment the training data with pose variations, keeping other hyperparameters consistent with the original paper. 
During testing, we evaluate scenarios where target observations with arbitrary poses are directly used as input. 
Additionally, we test scenarios where the inputs are transformed using an offline pose estimation method \cite{di2022gpv}, simulating the two-stage route of traditional methods with pre-canonicalizing (Uy \textit{et al}.\cite{JLcv21} + PE and U-RED \cite{URED} + PE).
For experiments on Scan2CAD, we directly use the baseline models trained on PartNet with the 25\% occlusion setting to conduct zero-shot testing, since real-world ground-truth models are inaccessible for training.

\textbf{Evaluation Metrics.}
We utilize Chamfer Distance (CD) on the magnitude of $10^{-2}$ to assess both full shape scenarios and partial shape scenarios. 
We calculate the metrics following \cite{JLcv21} to use the best result among the top 10 candidate objects. 
The final average metrics are obtained by averaging the results across all instances.

\textbf{Implementation Details.} 
During training, we uniformly sample objects to obtain point clouds with $M = 2500$ points to represent shapes.
We directly generate partial point clouds from the corresponding full point clouds by random cropping for the partial branch inputs. 
We apply random pose augmentation to the input shape, specifically with random translations $T_{rand} \in [-0.1,0.1]$ and random rotations $R_{rand} \in [-1,1]$ on three Eulerian angles respectively.
We set the initial learning rate to $1e-3$ and train ShapeMatcher for 200 epochs in every training stage of Sec. \ref{training}. 
Regarding the weight of the loss, in the first stage considering only the full branch, $\mathcal{L}_{can}$ and $\mathcal{L}_{seg}$ are equally weighted.
In the second stage introducing the partial branch, we primarily emphasize the partial-full consistency losses, assigning significant weights to $\mathcal{L}_{ccan}$, $\mathcal{L}_{ccen}$ and $\mathcal{L}_{cseg}$ with weights set as $5$, $2$, and $2$ respectively, while keeping the remaining weights default at $1$. 
In the final stage for joint $\mathbf{R\&D}$, both $\mathcal{L}_{retrieval}$ and $\mathcal{L}_{deform}$ are equally weighted.

\subsection{Synthetic Cases}
To validate the ability of ShapeMatcher tackling the challenge of arbitrary poses and occlusions, we first use synthetic datasets to simulate this scenario.
We evaluate all methods \cite{JLcv21,URED} where object observations with arbitrary poses are directly used as input.
Additionally, we also report results where inputs are transformed and canonicalized using an offline pose estimation method \cite{di2022gpv} for baseline methods (Uy \textit{et al}. + PE and U-RED + PE).
Moreover, we analyze inference time of ShapeMatcher against the $\mathbf{R\&D}$ baselines in the Supplementary Material. 

We conduct two types of inputs for evaluation: full inputs using the PartNet and ComplementMe datasets, and partial input tests using $10\%$, $25\%$, and $50\%$ occlusion rates on the PartNet dataset.
The results of the full input tests are detailed in Table \ref{tab:A}. 
In PartNet, our ShapeMatcher significantly outperforms the current leading competitors. 
For the Chamfer Distance on three categories, ShapeMatcher measures at $0.197$, $0.150$, and $0.519$, maintaining the leading position. 
Even when the processed PE results are used as input, the baselines' results still fall short of ShapeMatcher.
This demonstrates the effectiveness of adopting the affine-invariant features in the joint \textit{Canonicalization} step.
Results from ComplementMe supports the same conclusion, where ShapeMatcher reports significantly better results compared to the baseline methods. 
ShapeMatcher surpasses the top-performing Uy \textit{et al.} + PE by $85.2\%$.
Such superior results yielded by ShapeMatcher demonstrates that the joint consideration of all four steps improves the matching accuracy a lot.

For evaluation on partial inputs, we control the occlusion rates of partial point clouds by controlling the position of the cropping planes onto the full point clouds.
The evaluation on partial inputs are presented in Table \ref{tab:B}. 
Concretely, ShapeMatcher outperforms the current top method handling partial inputs U-RED by $5.018$, $5.666$ and $7.241$ at the occlusion rate of $10\%$, $25\%$ and $50\%$ respectively.
As the occlusion rate increases, the superiority of the ShapeMatcher method grows. 
Considering PE adopting, the same trend is exhibited.
ShapeMatcher surpasses the U-RED + PE by $0.525$, $0.949$, $2.434$ under three occlusion rates. 
It demonstrates that the proposed region-weighted retrieval brings strong robustness of our method against occlusion.

\begin{figure}[t]
   \centering
   \includegraphics[width=\linewidth]{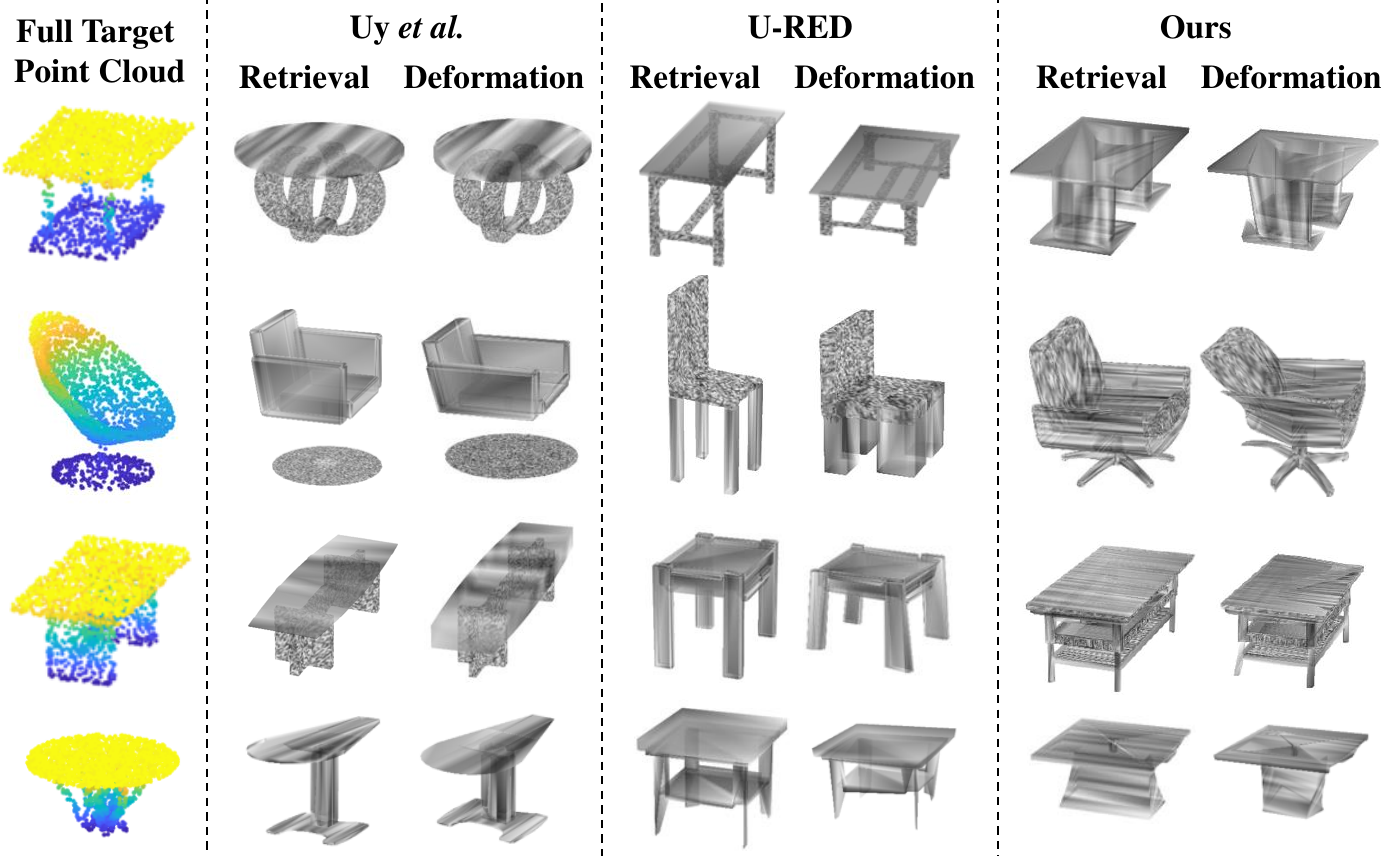}
   \caption{Qualitative \textbf{R\&D} results with full target inputs on PartNet.}
   \label{fig:ex_1}
 \end{figure}

\begin{figure}[t]
   \centering
   \includegraphics[width=\linewidth]{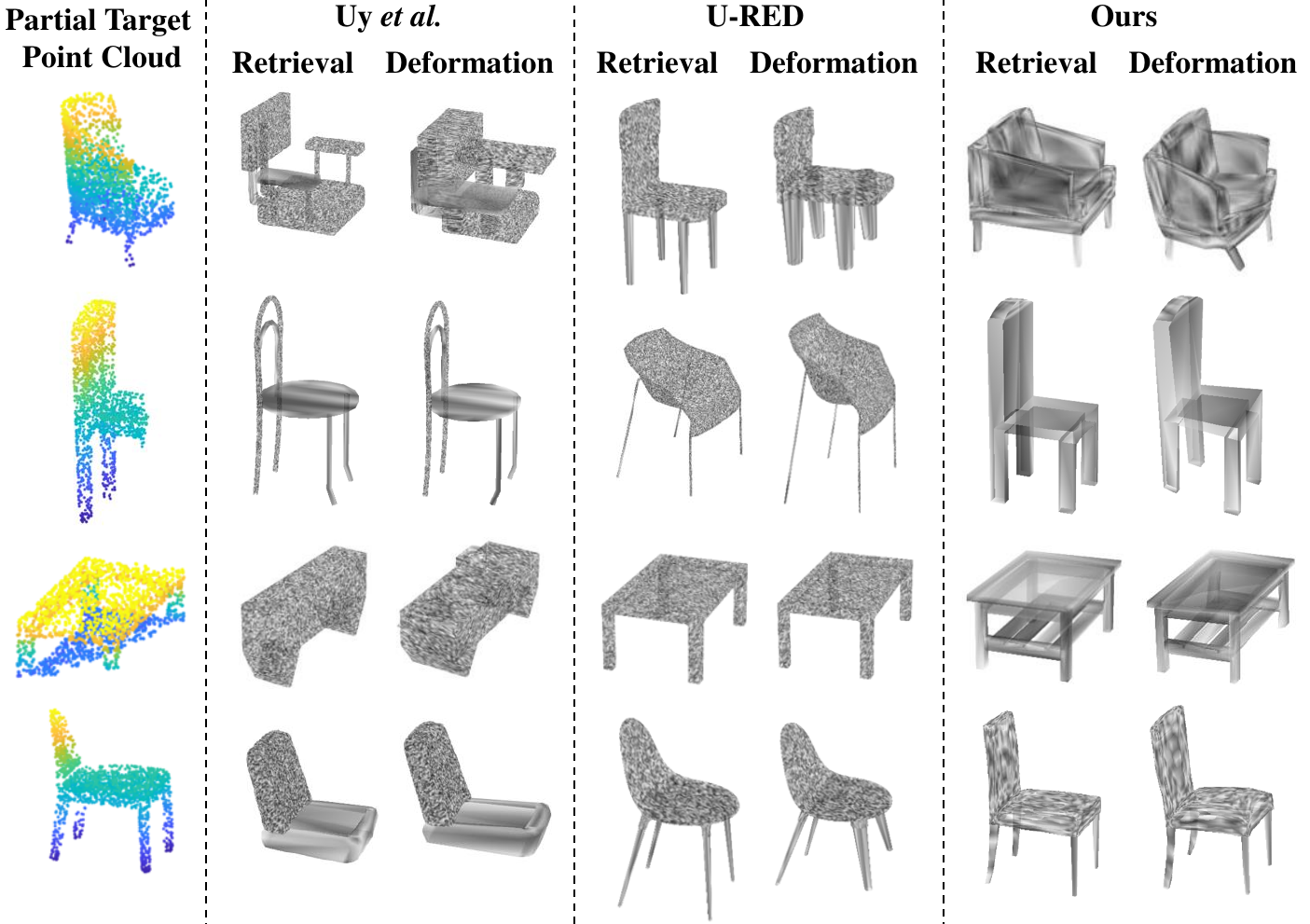}
   \caption{Qualitative \textbf{R\&D} results with partial target inputs on the occlusion rate of $25\%$ on PartNet.}
   \label{fig:ex_2}
 \end{figure}

As shown in Fig. \ref{fig:ex_1} and \ref{fig:ex_2}, our results shows more geometric resemblance to the targets compared to other methods. 
This is attributed to the suitable joint consideration of the four highly-associated processes, which accurately decouples the input poses, mapping them to a consistent space for accurate $R\&D$. 
The region-weighted retrieval we employ explicitly eliminates the influence of occluded areas, allowing for a more precise matching with the source model.

\begin{table}[!t]
  \centering
  \resizebox{\columnwidth}{!}{
  \begin{tabular}{ccccc}
    \hline
    \multicolumn{5}{c}{PartNet\cite{PNcv19}}\\
    \hline
    Method  & Chair & Table & Cabinet & Average\\
    \hline
    Uy \textit{et al.}\cite{JLcv21} & 4.269 & 6.302 & 4.118 & 5.271 \\
    Uy \textit{et al.}\cite{JLcv21} + PE & 1.507 & 3.006 & 1.070 & 2.219\\
    U-RED\cite{URED} & 5.331 & 4.980 & 9.141 & 5.463\\
    U-RED\cite{URED} + PE & 1.025 & 0.359 & 1.423 & 0.725\\
    Ours & \textbf{0.197} & \textbf{0.150} & \textbf{0.519} & \textbf{0.200}\\
    \hline
    \multicolumn{5}{c}{ComplementMe\cite{CMto17}}\\
    \hline
    Method  & Chair & Table & Cabinet & Average\\
    \hline
    Uy \textit{et al.}\cite{JLcv21} & 4.018 & 5.480 & -- & 4.825\\
    Uy \textit{et al.}\cite{JLcv21} + PE & 1.439 & 2.454 & -- & 1.999\\
    U-RED\cite{URED} & 8.575 & 5.800 & -- &7.044\\
    U-RED\cite{URED} + PE & 4.954 & 0.847 & -- &2.688\\
    Ours & \textbf{0.253} & \textbf{0.328} & -- &\textbf{0.294}\\
    \hline
  \end{tabular}
  }
  \caption{The Chamfer Distance metrics for joint \textbf{R$\&$D} results on full shapes under arbitrary poses.
   }
\label{tab:A}
\end{table}

\begin{table}[!t]
  \centering
  \resizebox{\columnwidth}{!}{
  \begin{tabular}{c|ccccc}
    \hline
    Occlusion & Method  & Chair & Table & Cabinet & Average\\
    \hline
    \multirow{5}{*}{10\%}
    & Uy \textit{et al.}\cite{JLcv21} & 4.372 & 6.395 & 4.179 & 5.365\\
    & Uy \textit{et al.} + PE\cite{JLcv21} & 1.523 & 2.982 & \textbf{1.133} & 2.219\\
    & U-RED\cite{URED} & 6.025 & 5.375 & 5.269 & 5.640\\
    & U-RED\cite{URED} + PE& 1.207 & 1.012 & 1.669 & 1.147\\
    & Ours & \textbf{0.676} & \textbf{0.481} & 1.212 & \textbf{0.622}\\
    \hline
    \multirow{5}{*}{25\%}
    & Uy \textit{et al.}\cite{JLcv21} & 4.654 & 6.927 & 4.750 & 5.795\\
    & Uy \textit{et al.} + PE\cite{JLcv21} & 1.803 & 3.195 & 1.607 & 2.481\\
    & U-RED\cite{URED} & 5.196 & 7.215 & 8.164 & 6.442\\
    & U-RED\cite{URED} + PE& 1.684 & 1.387 & 2.795 & 1.625\\
    & Ours & \textbf{0.878} & \textbf{0.643} & \textbf{1.071} & \textbf{0.776}\\
    \hline
    \multirow{5}{*}{50\%}
    & Uy \textit{et al.}\cite{JLcv21} & 6.070 & 9.322 & 7.929 & 7.841\\
    & Uy \textit{et al.} + PE\cite{JLcv21} & 3.314 & 5.030 & 4.584 & 4.272\\
    & U-RED\cite{URED} & 8.696 & 8.387 & 7.613 & 8.455\\
    & U-RED\cite{URED} + PE& 4.722 & 2.015 & 7.903 & 3.628\\
    & Ours & \textbf{1.197} & \textbf{1.079} & \textbf{1.872} & \textbf{1.194}\\
    \hline
  \end{tabular}
  }
  \caption{The Chamfer Distance metrics for joint \textbf{R$\&$D} results on partial shapes under arbitrary poses of PartNet dataset.
   }
\label{tab:B}
\end{table}

\subsection{Real-world Cases}
We test the effectiveness of ShapeMatcher on real-world datasets. 
In such case, ShapeMatcher is trained on the synthetic PartNet with $25\%$ occlusion and directly tested on the partial scans of the real-world dataset Scan2CAD without manual pose adjustments. 
Table \ref{tab:C} displays our results, where our method significantly outperforms existing competitors. 
Particularly, compared to the U-RED, in three categories, the reported Chamfer Distance are reduced by $92\%$, $96\%$, and $94\%$ respectively. 
In comparison to Uy \textit{et al.}, the reported Chamfer Distance are reduced by $91\%$, $98\%$, and $92\%$. 
Even compared with the use of PE on the two baseline methods, there remains a significant leap forward of ShapeMatcher.
This zero-shot experiments on real-world further prove that the procedure of joint \textit{Canonicalization}, \textit{Segmentation}, \textit{Retrieval} and \textit{Deformation} possesses a strong domain adaptation ability, showcasing the great potential of ShapeMatcher in real-world applications. 
The visualizations on Scan2CAD are provided in the supplementary material.

\begin{table}[!t]
  \centering
  \resizebox{\columnwidth}{!}{
  \begin{tabular}{ccccc}
    \hline
    Method  & Chair & Table & Cabinet & Average\\
    \hline
    Uy \textit{et al.}\cite{JLcv21} & 4.886 & 7.605 & 8.335 &  6.181 \\
    Uy \textit{et al.}\cite{JLcv21} + PE & 3.362 & 6.657  & 7.261  & 4.905\\
    U-RED\cite{URED} & 5.490 & 5.131 & 10.091 & 5.945\\
    U-RED\cite{URED} + PE &2.893  &3.164  &5.957  &3.354\\
    Ours & \textbf{0.423} & \textbf{0.186} & \textbf{0.654} &\textbf{0.375}\\
    \hline
  \end{tabular}
  }
  \caption{The Chamfer Distance metrics for joint \textbf{R$\&$D} results in real-world Scan2CAD \cite{S2C}. 
   }
\label{tab:C}
\end{table}

\subsection{Ablation Studies}
We conduct ablation experiments on PartNet, mainly on two aspects. 
First, in the \textit{Canonicalization}, we investigate the importance of disentangling different pose intrinsics in Table \ref{tab:D}, and demonstrate the effectiveness of joint considering \textit{Canonicalization}.
Second, in Table \ref{tab:E}, we ablate the region-weighted \textit{Retrieval} and the part center guided \textit{Deformation}.
Moreover, the robustness against occlusion is analyzed in Table \ref{tab:B}.

\textbf{Canonicalization Capability.}
To study the impact of different intrinsics of object poses in the \textit{Canonicalization} process, we conduct ablations on decoupling of translations, rotations and scales. The results are presented in Table \ref{tab:D}. 
Specifically, in row (1), we make no adjustments to the input poses. 
Thanks to the regional-level $R\&D$ process, it shows decent performance. 
However, there is still a noticeable gap compared to row (4), indicating the significance of our proposed joint \textit{Canonicalization}. 
In row (2), we solely decouple the input translations, resulting in a decrease of $14\%$ in reported metrics. 
Moving to row (3), upon this foundation, we add decoupling for rotation, leading to a substantial decrease in reported Chamfer Distance, averaging at $0.244$. 
In row (4), we introduce scale decoupling, resulting in another decrease in the reported metrics. 
It is evident that accurate decoupling of rotation is a crucial aspect of the success of the \textit{Canonicalization} process.
Moreover, it demonstrates that to integrate \textit{Canonicalization} and $R\&D$ process is indispensable for the ShapeMatcher process.

\textbf{Deformation and Retrieval Ability.}
To validate the effectiveness of our proposed region-weighted \textit{Retrieval} and the part center guided neural cage \textit{Deformation}, we conduct an ablation study on the PartNet dataset with the $25\%$ occlusion rate.
The results are presented in Table \ref{tab:E}.
In row (1), we conduct experiments using global retrieval and global deformation. 
This means we directly use an MLP network to extract overall point cloud features as the retrieval vector \cite{JLcv21}. 
In the deformation network, similarly, we directly use an MLP network to generate neural cage offsets for deformation \cite{NCcv20,KPcv21}. 
Due to the lack of extraction of local information, the reported Chamfer Distance is more than twice of row (4).
In row (2), we employ the global retrieval and the part center guided neural cage deformation. 
This improvement allows much more tightly-matched deformation by the retrieved source model, resulting in a $14\%$ decrease in reported metrics.
In row (3), we conduct experiments using the regional retrieval and the global deformation. 
The proposed regional-weighted retrieval handles occluded objects, reducing the impact of occluded parts and resulting in a substantial decrease in Chamfer Distance, down to $0.973$.

\textbf{Occlusion Robustness.}
We test shapes at different occlusion levels by altering the occlusion ratio in the input. 
For each specific occlusion ratio, we deliberately crop a portion of the complete point cloud to simulate occlusion. 
We test scenarios with occlusion ratios of $10\%$, $25\%$, and $50\%$, and the results are presented in Table \ref{tab:B}.
Observably, as the occluded regions increased, the Chamfer Distance significantly rises for the baseline methods. 
Taking the U-RED + PE as an example, its reported metrics increase from $1.147$ to $3.628$, doubling in value, as the occluded area expands. 
In contrast, our method increases by less than $1$-fold, which exhibits strong robustness against occlusion.

\begin{table}[!t]
  \centering
  \resizebox{\columnwidth}{!}{
  \begin{tabular}{c|ccc|cccc}
    \hline
      &Trans.  & Rot. & Scal. & Chair & Table & Cabinet & Average\\
    \hline
    (1) & & & & 0.571 & 0.502 & 1.233 & 0.590\\
    (2) &\checkmark & & & 0.468 & 0.442 & 1.096 & 0.506\\
    (3) &\checkmark & \checkmark & & 0.213 & 0.200 & 0.674 & 0.244\\
    (4) & \checkmark & \checkmark & \checkmark & \textbf{0.197} &\textbf{0.150} &\textbf{0.519}  &\textbf{0.200}\\
    \hline
  \end{tabular}
  }
  \caption{Ablations on the \textit{Canonicalization} process, we demonstrate the effectiveness of the joint \textit{Canonicalization} by ablating different pose intrinsics. 
  Here, Trans. denotes the decoupling of translation, Rot. represents the rotation, and Scal. signifies the scale.}
\label{tab:D}
\end{table}

\begin{table}[!t]
  \centering
  \resizebox{\columnwidth}{!}{
  \begin{tabular}{c|cccc|cccc}
    \hline
      & Gl. R. & Re. R. & Gl. D. & Re. D. & Chair & Table & Cabinet & Average\\
    \hline
    (1) & \checkmark & &\checkmark & & 1.672 & 1.446 & 1.800 & 1.570\\
    (2) &\checkmark &  & &\checkmark & 1.539 & 1.305 & 1.641 & 1.431\\
    (3) & & \checkmark & \checkmark & & 1.042 & 0.874 & 1.223 & 0.973\\
    (4) & & \checkmark & & \checkmark &\textbf{0.878} &\textbf{0.643} &\textbf{1.071} & \textbf{0.776}\\
    \hline
  \end{tabular}
  }
  \caption{Ablations of the $R\&D$ process.
  GL. R. denotes the global feature based retrieval \cite{JLcv21}, Re. R. represents the proposed region-weighted retrieval, GL. D. signifies direct neural cage deformation using global features \cite{NCcv20,KPcv21}, and Re. D. denotes the adopted regional part center guided neural cage deformation.}
\vspace{-0.5cm}
\label{tab:E}
\end{table}
\section{Conclusion}
In this paper, we present ShapeMatcher, a unified self-supervised learning framework for joint shape canonicalization, segmentation, retrieval and deformation.
Given a partially-observed object in an arbitrary pose, we first canonicalize the object by extracting point-wise affine-invariant features.
Then, the affine-invariant features are leveraged to predict semantically consistent part segmentation and corresponding part centers.
Afterwards, the lightweight region-weighted retrieval module aggregates the features within each part as its retrieval token and compare all the tokens with source shapes from a pre-established database to identify the most geometrically similar shape.
Finally, we deform the retrieved shape in the deformation module to tightly fit the input object by harnessing part center guided neural cage deformation.
Extensive experiments on synthetic datasets PartNet, ComplementMe, and real-world dataset Scan2CAD demonstrate that ShapeMatcher surpasses competitors by a large margin.
In the future, we plan to further applicate I-RED to various downstream tasks like robotic grasping.
\textbf{Limitations.} are discussed in the Supplementary Material.

{\small
\bibliographystyle{ieee_fullname}
\bibliography{egbib}
}

\end{document}